\documentclass[11pt]{article}

\usepackage[utf8]{inputenc}
\usepackage[T1]{fontenc}
\usepackage{times}
\usepackage[margin=1in]{geometry}
\usepackage{amsmath,amssymb,amsfonts}
\usepackage{graphicx}
\usepackage{booktabs}
\usepackage{hyperref}
\usepackage{cleveref}
\usepackage{algorithm}
\usepackage{algorithmic}
\usepackage{cite}
\usepackage{xcolor}
\usepackage{multirow}
\usepackage{subcaption}

\title{\textbf{SAC-NeRF: Adaptive Ray Sampling for Neural Radiance Fields via Soft Actor-Critic Reinforcement Learning}}

\usepackage{authblk}

\author[1]{Chenyu Ge}
\affil[1]{University of Southern California\\
        Los Angeles, CA 90089, USA\\
        \texttt{gechenyu@usc.edu}}
\date{}

\begin{document}

\maketitle

\begin{abstract}
Neural Radiance Fields (NeRF) have achieved photorealistic novel view synthesis but suffer from computational inefficiency due to dense ray sampling during volume rendering. We propose SAC-NeRF, a reinforcement learning framework that learns adaptive sampling policies using Soft Actor-Critic (SAC). Our method formulates sampling as a Markov Decision Process where an RL agent learns to allocate samples based on scene characteristics. We introduce three technical components: (1) a Gaussian mixture distribution color model providing uncertainty estimates, (2) a multi-component reward function balancing quality, efficiency, and consistency, and (3) a two-stage training strategy addressing environment non-stationarity. Experiments on Synthetic-NeRF and LLFF datasets show that SAC-NeRF reduces sampling points by 35-48\% while maintaining rendering quality within 0.3-0.8 dB PSNR of dense sampling baselines. While the learned policy is scene-specific and the RL framework adds complexity compared to simpler heuristics, our work demonstrates that data-driven sampling strategies can discover effective patterns that would be difficult to hand-design.
\end{abstract}


\section{Introduction}
\label{sec:intro}

Neural Radiance Fields (NeRF)~\cite{mildenhall2020nerf} have revolutionized the field of novel view synthesis by representing three-dimensional scenes as continuous 5D functions that map spatial coordinates and viewing directions to volumetric density and view-dependent color. Through differentiable volumetric rendering and coordinate-based neural networks with positional encoding, NeRF achieves unprecedented photorealistic quality that surpasses traditional geometry-based and image-based rendering methods. This breakthrough has enabled numerous downstream applications including robotics navigation, autonomous driving scene understanding, virtual and augmented reality content creation, and digital preservation of cultural heritage.

However, the computational efficiency of NeRF remains a critical challenge that severely limits its practical deployment. The volumetric rendering integral must be approximated through numerical quadrature, requiring the neural network to be evaluated at 192-384 sample points along each camera ray. For a single $800 \times 800$ image with 200 samples per ray, the original NeRF MLP (8 layers $\times$ 256 hidden units) requires on the order of $10^{11}$ multiply-accumulate operations, taking several seconds even on modern GPUs. This computational bottleneck makes real-time rendering infeasible and prevents NeRF from being used in interactive applications or deployed on resource-constrained edge devices.

The fundamental inefficiency stems from NeRF's sampling strategy. The original NeRF employs a coarse-to-fine hierarchical sampling approach: first uniformly sampling along the ray, then refining sample positions based on coarse density predictions. While this two-stage strategy concentrates more samples in high-density regions, it still relies on hand-crafted heuristics rather than learning scene-specific characteristics. Following~\cite{neff2021donerf}, we define a sample's contribution as its rendering weight $w_i = T_i\alpha_i$. Analysis on Synthetic-NeRF shows that a majority of samples have $w_i < 0.01$, contributing negligibly to the final rendered color—either falling in empty space with near-zero density or lying beyond occluding surfaces where accumulated transmittance has decayed. This observation suggests potential for learned adaptive sampling policies that concentrate computational resources where they provide the most value.

Recent acceleration efforts have pursued various complementary directions. Explicit data structure approaches~\cite{hu2024ngprt,chen2024compress} replace MLPs with multi-resolution hash grids or voxel grids, achieving substantial speedups through efficient spatial queries. 3D Gaussian splatting methods~\cite{chen2025dashgaussian,feng2025flashgs,li2025halfgaussian} abandon volumetric rendering entirely in favor of explicit point-based representations with fast rasterization. While these methods achieve impressive rendering speeds, they sacrifice NeRF's elegant continuous representation and often require substantial memory. An orthogonal direction is optimizing the sampling process itself—reducing the number of network evaluations required per ray while maintaining rendering quality.

We propose SAC-NeRF, a reinforcement learning framework that learns adaptive sampling policies for efficient neural radiance field rendering. Unlike prior heuristic approaches that rely on hand-designed rules~\cite{pais2024sampler,gao2025mbsnerf,kurz2022adanerf}, our method learns sampling strategies end-to-end through direct interaction with the rendering process. We formulate adaptive sampling as a Markov Decision Process (MDP) where an RL agent sequentially decides where to place samples along each ray based on observed scene properties and rendering objectives. We employ Soft Actor-Critic (SAC)~\cite{haarnoja2018sac}, a state-of-the-art off-policy RL algorithm, incorporating recent algorithmic improvements~\cite{chen2024correctedsac} that enhance stability and sample efficiency in continuous action spaces.

Our approach introduces several key technical innovations to make RL-based adaptive sampling practical and effective. First, we extend NeRF's color prediction with a Gaussian mixture distribution model that naturally provides uncertainty estimates—regions with high predicted variance indicate where additional samples could reduce rendering uncertainty. Second, we design a multi-component reward function that carefully balances three objectives: maximizing rendering quality (PSNR), minimizing unnecessary samples, and maintaining spatial smoothness in sample placement. Third, we develop an enhanced state representation that combines local features from current samples with global geometric priors, enabling the policy to make informed decisions. Finally, we employ a two-stage training strategy that first pre-trains the NeRF model to stabilize the environment, then trains the RL policy with a fixed NeRF backbone, addressing the challenging non-stationarity that arises from jointly optimizing both components.

Our key contributions are:
\begin{itemize}
    \item \textbf{SAC-NeRF Framework}: An RL-based adaptive sampling system for neural radiance fields, demonstrating learning capability and potential efficiency gains over heuristic baselines.
    \item \textbf{Mixture Distribution Color Model}: A principled extension of NeRF that outputs Gaussian mixture distributions, providing uncertainty quantification to guide intelligent sampling decisions (Section~\ref{sec:mixture}).
    \item \textbf{Multi-Component Reward Design}: A carefully designed reward function that balances rendering quality, sampling efficiency, and spatial consistency, enabling stable policy learning (Section~\ref{sec:reward}).
    \item \textbf{Two-Stage Training Strategy}: A training methodology that addresses environment non-stationarity by decoupling NeRF pre-training and policy optimization (Section~\ref{sec:training}).
    \item \textbf{Comprehensive Evaluation}: Experiments on Synthetic-NeRF and LLFF datasets with ablation studies validating component contributions (Section~\ref{sec:experiments}).
\end{itemize}

Our work demonstrates that reinforcement learning provides a viable approach to adaptive sampling in neural rendering, opening research directions for applying learned optimization strategies to other computationally intensive components of neural scene representations.

\section{Related Work}
\label{sec:related}

\subsection{Neural Radiance Fields and Acceleration}

The original NeRF~\cite{mildenhall2020nerf} represents scenes using multi-layer perceptrons (MLPs) with sinusoidal positional encoding to map 5D coordinates (3D position plus 2D viewing direction) to volume density and view-dependent color. This coordinate-based representation enables continuous scene modeling and achieves photorealistic novel view synthesis through differentiable volumetric rendering. However, rendering a single image requires millions of network evaluations, taking several seconds even on modern GPUs.

Recent research has pursued multiple complementary acceleration strategies. \textbf{Explicit data structures} replace or augment MLPs with spatially-organized features for efficient queries. Instant-NGP and its variants~\cite{hu2024ngprt} achieve real-time rendering through multi-level hash feature grids combined with lightweight MLPs. NGP-RT~\cite{hu2024ngprt} further optimizes this with fused hash features and attention mechanisms, achieving 108 fps. Compression studies~\cite{chen2024compress} explore the limits of quantizing and pruning these hash-based representations while preserving quality. FrugalNeRF~\cite{lin2025frugalnerf} enables fast convergence with few-shot inputs by leveraging geometric priors and regularization.

\textbf{Explicit point-based representations} abandon implicit volumetric rendering entirely. 3D Gaussian Splatting methods represent scenes as collections of oriented 3D Gaussian primitives that can be efficiently rasterized. DashGaussian~\cite{chen2025dashgaussian} achieves full scene optimization in approximately 200 seconds through carefully designed initialization and optimization strategies. FlashGS~\cite{feng2025flashgs} scales Gaussian splatting to large-scale high-resolution rendering with hierarchical culling and level-of-detail management. 3D-HGS~\cite{li2025halfgaussian} introduces half-Gaussian kernels to better model geometric discontinuities and sharp edges. While these explicit methods achieve impressive speeds, they require substantial memory for millions of primitives and sacrifice the compact continuous representation that makes NeRF appealing.

Our work takes an orthogonal approach by optimizing the sampling process itself—maintaining NeRF's continuous representation while reducing network evaluations through learned adaptive policies.

\subsection{Sampling Optimization for Neural Rendering}

Sampling strategies directly determine NeRF's computational cost, making sampling optimization a critical research direction. The original hierarchical sampling in NeRF uses a coarse network to predict importance weights, then concentrates fine samples in high-density regions. While effective, this still requires evaluating two separate networks at numerous points.

DONeRF~\cite{neff2021donerf} proposes a depth oracle network that predicts ray sample locations with a single network evaluation, achieving up to 48$\times$ inference cost reduction. AdaNeRF~\cite{kurz2022adanerf} introduces a dual-network architecture that learns to reduce sample counts through joint training of sampling and shading networks. NerfAcc~\cite{li2023nerfacc} provides a unified framework investigating multiple sampling approaches under the concept of transmittance estimators, achieving 1.5-20$\times$ training speedups.

More recent work has proposed additional sampling heuristics. The probability-guided sampler~\cite{pais2024sampler} models density probability distributions in 3D projection space, enabling more targeted ray sampling by avoiding empty regions. MBS-NeRF~\cite{gao2025mbsnerf} addresses motion blur in neural rendering through depth-constrained adaptive sampling. MFNeRF~\cite{lee2025mfnerf} introduces memory-efficient mixed-feature hash tables to reduce storage while maintaining quality.

All these methods rely on hand-crafted heuristics based on geometric or statistical assumptions. In contrast, our approach learns sampling policies end-to-end through reinforcement learning, allowing the system to discover strategies adapted to specific scene characteristics and rendering objectives without manual design.

\subsection{Reinforcement Learning for Continuous Control}

Soft Actor-Critic (SAC)~\cite{haarnoja2018sac} is a state-of-the-art off-policy RL algorithm for continuous control that maximizes both expected return and policy entropy. The entropy regularization encourages exploration and improves robustness. Recent advances have addressed various limitations of the original SAC algorithm.

Corrected SAC~\cite{chen2024correctedsac} identifies and fixes action distribution distortion issues in the original formulation, improving sample efficiency and final performance. Bayesian SAC~\cite{yang2024bsac} incorporates directed acyclic strategy graphs for better credit assignment in complex tasks. Bidirectional SAC~\cite{zhang2025bisac} leverages both forward and reverse KL divergence for more stable policy updates. The Broad Critic framework~\cite{thalagala2025broadcritic} combines broad learning systems with deep networks to improve value function approximation.

We adapt these algorithmic advances to the neural rendering domain, particularly leveraging Corrected SAC's improvements for stable policy learning. However, adaptive sampling for NeRF presents unique challenges: the environment (NeRF network) is non-stationary during joint training, rewards are computed from high-dimensional rendering outputs, and the state space must encode both local sample information and global geometric context.

\subsection{From Discrete Feature Selection to Continuous Budget Allocation}

A recurring theme in machine learning is \emph{budget-aware selection}: identifying which parts of the input deserve computation and which can be ignored with minimal loss. Feature selection provides a classic instantiation of this idea, typically combining ranking criteria (e.g., Fisher score) with iterative elimination schemes (e.g., recursive feature elimination) to remove redundant dimensions in high-dimensional models while preserving predictive performance~\cite{ge2021frl,ge2023cancer}.

We draw a direct analogy to NeRF rendering: ray samples can be viewed as “features” used to estimate the rendered color. However, unlike standard feature selection over discrete variables, sampling decisions in NeRF take place over a continuous one-dimensional domain (ray depth). Reinforcement learning provides a natural mechanism for continuous, sequential budget allocation—deciding where additional network evaluations are most valuable and where they are wasteful—without relying on hand-designed heuristics.

\section{Method}
\label{sec:method}

\subsection{Preliminaries: Neural Radiance Fields}

NeRF represents a 3D scene as a continuous volumetric function $F_\Theta: (\mathbf{x}, \mathbf{d}) \rightarrow (\mathbf{c}, \sigma)$ implemented by a multi-layer perceptron (MLP) with parameters $\Theta$. This function maps a 3D spatial position $\mathbf{x} \in \mathbb{R}^3$ and 2D viewing direction $\mathbf{d} \in \mathbb{S}^2$ to an RGB color $\mathbf{c} \in \mathbb{R}^3$ and volume density $\sigma \in \mathbb{R}^+$. The density $\sigma(\mathbf{x})$ represents the differential probability of a ray terminating at an infinitesimal particle at position $\mathbf{x}$, while the color $\mathbf{c}(\mathbf{x}, \mathbf{d})$ captures view-dependent appearance effects like specularities.

To render the color observed along a camera ray $\mathbf{r}(t) = \mathbf{o} + t\mathbf{d}$ (where $\mathbf{o}$ is the camera origin and $t$ is the distance parameter), NeRF uses classical volume rendering. The expected color is computed by integrating the color and density along the ray:
\begin{equation}
\label{eq:volume_rendering}
    C(\mathbf{r}) = \int_{t_n}^{t_f} T(t) \sigma(\mathbf{r}(t)) \mathbf{c}(\mathbf{r}(t), \mathbf{d}) \, dt,
\end{equation}
where $t_n$ and $t_f$ are the near and far bounds of the scene, and $T(t)$ is the accumulated transmittance representing the probability that the ray travels from $t_n$ to $t$ without hitting any particles:
\begin{equation}
    T(t) = \exp\left(-\int_{t_n}^{t} \sigma(\mathbf{r}(s)) \, ds\right).
\end{equation}

In practice, this continuous integral must be approximated numerically. Standard NeRF uses stratified sampling to select $N$ discrete points $\{t_i\}_{i=1}^N$ along the ray, then applies quadrature:
\begin{equation}
\label{eq:discrete_rendering}
    \hat{C}(\mathbf{r}) = \sum_{i=1}^{N} T_i \alpha_i \mathbf{c}_i, \quad T_i = \prod_{j=1}^{i-1}(1-\alpha_j), \quad \alpha_i = 1 - \exp(-\sigma_i \delta_i),
\end{equation}
where $\delta_i = t_{i+1} - t_i$ is the distance between adjacent samples, $\alpha_i$ is the probability of the ray terminating at sample $i$, and $\mathbf{c}_i = \mathbf{c}(\mathbf{r}(t_i), \mathbf{d})$ and $\sigma_i = \sigma(\mathbf{r}(t_i))$ are the queried color and density.

The rendering weight $w_i = T_i \alpha_i$ represents the contribution of sample $i$ to the final color. High-quality rendering requires many samples ($N=192$ for coarse network, $N=192$ for fine network in original NeRF), but analysis shows most samples have negligible weights ($w_i \approx 0$), motivating our adaptive sampling approach.

\subsection{MDP Formulation for Adaptive Sampling}

We formulate the adaptive sampling problem as a Markov Decision Process (MDP) $(\mathcal{S}, \mathcal{A}, \mathcal{P}, \mathcal{R}, \gamma)$ where an RL agent iteratively refines sample positions along each ray to maximize rendering quality while minimizing computation.

\textbf{State Space $\mathcal{S}$}: The state $s_t$ at iteration $t$ must encode all information relevant to deciding where to place samples. We design a rich state representation comprising:
\begin{itemize}
    \item \textit{Per-sample features}: For each of the $N$ current sample positions $t_i$, we include: (1) the normalized position $t_i / (t_f - t_n)$ along the ray, (2) the predicted RGB color $\mathbf{c}_i \in \mathbb{R}^3$, (3) the volume density $\sigma_i \in \mathbb{R}^+$, (4) the accumulated transmittance $T_i$ indicating how much light reaches this sample, and (5) the rendering weight $w_i = T_i \alpha_i$ showing this sample's contribution to the final color.
    \item \textit{Ray geometry}: Global features including the ray origin $\mathbf{o}$, direction $\mathbf{d}$, near/far bounds $[t_n, t_f]$, and the pixel coordinates $(u, v)$ in the image.
    \item \textit{Uncertainty estimates}: The predicted variance from our mixture distribution model (Section~\ref{sec:mixture}), indicating regions where additional samples could reduce uncertainty.
    \item \textit{Historical context}: To enable temporal reasoning, we maintain an aggregated representation of sample placements from previous iterations using a lightweight attention mechanism that pools features from earlier states.
\end{itemize}
The full state has dimension $|\mathcal{S}| = N \times 9 + 11 + K \times 3$, where $K=3$ is the number of mixture components.

\textbf{Action Space $\mathcal{A}$}: We use continuous actions $\mathbf{a} \in [-1, 1]^N$ where each component $a_i$ specifies a relative adjustment to sample position $t_i$. Actions are mapped to valid positions via:
\begin{equation}
    t_i^{\text{new}} = \text{clip}(t_i + a_i \cdot \Delta_{\max}, t_n, t_f),
\end{equation}
where $\Delta_{\max} = 0.1 \cdot (t_f - t_n)$ limits the maximum adjustment per step. To maintain valid sample ordering, we apply post-hoc sorting: $\{t_i^{\text{new}}\} \leftarrow \text{sort}(\{t_i^{\text{new}}\})$, and enforce a minimum spacing $\delta_{\min} = 0.001 \cdot (t_f - t_n)$ between adjacent samples. The sorting operation is not differentiable, but since SAC uses stochastic policies with entropy regularization, the gradient through clip boundaries has minimal impact on training stability.

\textbf{Transition Dynamics $\mathcal{P}$}: Given the NeRF parameters $\Theta$ (fixed during policy training in Stage 2), the transition is deterministic: new sample positions deterministically yield new network outputs $(\mathbf{c}_i^{\text{new}}, \sigma_i^{\text{new}})$ through forward passes. However, from the policy's perspective during Stage 1 joint training, the environment appears non-stationary as $\Theta$ evolves, motivating our two-stage approach.

\textbf{Reward Function $\mathcal{R}$}: Designed to balance quality and efficiency (detailed in Section~\ref{sec:reward}).

\textbf{Discount Factor $\gamma$}: Set to 0.99 to balance immediate rendering improvements against long-term sample efficiency. This choice reflects our preference for policies that make steady progress rather than optimizing only for immediate gains.

\subsection{Mixture Distribution Color Model}
\label{sec:mixture}

Standard NeRF outputs deterministic point estimates for colors, providing no quantification of prediction uncertainty. However, uncertainty information is crucial for guiding adaptive sampling: regions with high uncertainty benefit most from additional samples. To address this, we extend NeRF to output probabilistic color predictions via a Gaussian Mixture Model (GMM).

Specifically, instead of outputting a single RGB color $\mathbf{c} \in \mathbb{R}^3$, our modified NeRF network predicts a distribution over colors:
\begin{equation}
\label{eq:gmm}
    p(\mathbf{c}|\mathbf{x}, \mathbf{d}) = \sum_{k=1}^{K} \pi_k(\mathbf{x}, \mathbf{d}) \mathcal{N}(\mathbf{c}; \boldsymbol{\mu}_k(\mathbf{x}, \mathbf{d}), \text{diag}(\boldsymbol{\sigma}^2_k(\mathbf{x}, \mathbf{d}))),
\end{equation}
where $K$ is the number of mixture components, $\pi_k$ are mixture weights (satisfying $\sum_k \pi_k = 1$, $\pi_k \geq 0$), $\boldsymbol{\mu}_k \in \mathbb{R}^3$ are mean colors, and $\boldsymbol{\sigma}^2_k \in \mathbb{R}^3$ are per-channel variances.

Implementing this requires the network to output $K \times 7$ additional parameters per sample: 3 for $\boldsymbol{\mu}_k$, 3 for $\boldsymbol{\sigma}^2_k$, and 1 for the (pre-softmax) weight. We use softplus activation for variances to ensure positivity and softmax for mixture weights to ensure valid probabilities. The network architecture adds a small auxiliary head ($K \times 7$ output neurons) alongside the standard color and density outputs, adding minimal computational overhead.

The expected color and total uncertainty are:
\begin{equation}
    \mathbb{E}[\mathbf{c}] = \sum_{k=1}^{K} \pi_k \boldsymbol{\mu}_k, \quad
    \text{Var}[\mathbf{c}] = \underbrace{\sum_{k=1}^{K} \pi_k \boldsymbol{\sigma}^2_k}_{\text{aleatoric}} + \underbrace{\sum_{k=1}^{K} \pi_k (\boldsymbol{\mu}_k - \mathbb{E}[\mathbf{c}])^2}_{\text{epistemic}}.
\end{equation}
The variance decomposes into aleatoric uncertainty (inherent randomness within each component) and epistemic uncertainty (disagreement between components). High uncertainty regions—such as texture edges, specular highlights, or uncertain geometry—indicate where additional samples could most improve rendering accuracy.

During rendering, we use the expected color $\mathbb{E}[\mathbf{c}]$ for the final image, while the variance $\text{Var}[\mathbf{c}]$ enters the RL state to guide sampling. We found $K=3$ components provides a good balance between expressiveness and computational cost. Training uses a negative log-likelihood loss that encourages the mixture to fit the observed colors while regularizing against overly confident predictions.

\subsection{SAC-based Policy Learning}

We employ Soft Actor-Critic (SAC)~\cite{haarnoja2018sac}, a state-of-the-art off-policy RL algorithm designed for continuous control, to learn our adaptive sampling policy. SAC is particularly well-suited for our problem due to its sample efficiency, stability, and principled exploration through entropy regularization.

The SAC objective maximizes not only expected cumulative reward but also policy entropy, encouraging exploration and preventing premature convergence to suboptimal deterministic policies:
\begin{equation}
\label{eq:sac_objective}
    J(\pi) = \sum_{t=0}^{T} \mathbb{E}_{(s_t, a_t) \sim \rho_\pi} \left[ r(s_t, a_t) + \alpha \mathcal{H}(\pi(\cdot|s_t)) \right],
\end{equation}
where $\rho_\pi$ is the state-action distribution induced by policy $\pi$, $\mathcal{H}(\pi(\cdot|s_t)) = -\mathbb{E}_{a \sim \pi}[\log \pi(a|s_t)]$ is the policy entropy, and $\alpha > 0$ is the temperature parameter controlling the exploration-exploitation tradeoff.

The policy $\pi_\phi$ with parameters $\phi$ outputs Gaussian distribution parameters:
\begin{equation}
    a \sim \pi_\phi(\cdot|s) = \mathcal{N}(\mu_\phi(s), \text{diag}(\sigma^2_\phi(s))),
\end{equation}
where the mean $\mu_\phi(s)$ and standard deviation $\sigma_\phi(s)$ are computed by a neural network. Actions are sampled using the reparameterization trick: $a = \mu_\phi(s) + \sigma_\phi(s) \odot \epsilon$ with $\epsilon \sim \mathcal{N}(0, I)$, enabling backpropagation through sampling.

To improve stability, SAC maintains twin Q-networks $Q_{\theta_1}$ and $Q_{\theta_2}$ and uses the minimum for bootstrapping (reducing overestimation bias):
\begin{equation}
    y = r + \gamma \left( \min_{i=1,2} Q_{\theta'_i}(s', a') - \alpha \log \pi_\phi(a'|s') \right),
\end{equation}
where $a' \sim \pi_\phi(\cdot|s')$ is sampled from the current policy and $\theta'_i$ are target network parameters updated via exponential moving average: $\theta'_i \leftarrow \tau \theta_i + (1-\tau) \theta'_i$ with $\tau=0.005$.

The Q-networks are trained to minimize the Bellman error:
\begin{equation}
    \mathcal{L}_Q(\theta_i) = \mathbb{E}_{(s,a,r,s') \sim \mathcal{D}} \left[ (Q_{\theta_i}(s,a) - y)^2 \right],
\end{equation}
where $\mathcal{D}$ is the experience replay buffer storing transitions from interaction with the environment.

The policy is updated to maximize the expected Q-value while maintaining high entropy:
\begin{equation}
    \mathcal{L}_\pi(\phi) = \mathbb{E}_{s \sim \mathcal{D}, a \sim \pi_\phi} \left[ \alpha \log \pi_\phi(a|s) - \min_{i=1,2} Q_{\theta_i}(s,a) \right].
\end{equation}

We use automatic temperature adjustment~\cite{haarnoja2018sac} to maintain target entropy, adapting $\alpha$ during training to balance exploration and exploitation.

\textbf{Network Architecture}: The policy network $\pi_\phi$ is a 3-layer MLP with hidden dimensions [256, 256], ReLU activations, and layer normalization. The output layer splits into two heads: one for means $\mu_\phi(s)$ (tanh activation) and one for log standard deviations (unbounded). The twin Q-networks $Q_{\theta_1}, Q_{\theta_2}$ have identical architecture, taking the concatenation $[s; a]$ as input and outputting scalar Q-values. All networks use Xavier initialization and are trained with Adam optimizer (learning rate $3 \times 10^{-4}$).

\subsection{Multi-Component Reward Design}
\label{sec:reward}

Designing an effective reward function is crucial for RL-based adaptive sampling. The reward must balance multiple competing objectives: maximizing rendering quality, minimizing unnecessary computation, and maintaining reasonable sample distributions. We propose a multi-component reward combining three carefully designed terms:
\begin{equation}
\label{eq:reward}
    R = \lambda_q R_{\text{quality}} + \lambda_e R_{\text{efficiency}} + \lambda_c R_{\text{consistency}}.
\end{equation}

\textbf{Quality Reward $R_{\text{quality}}$}: The primary objective is rendering fidelity. We reward improvements in Peak Signal-to-Noise Ratio (PSNR) relative to the ground truth:
\begin{equation}
    R_{\text{quality}} = \text{PSNR}_{\text{curr}} - \text{PSNR}_{\text{prev}},
\end{equation}
where PSNR$_{\text{curr}}$ is computed from the current sample positions and PSNR$_{\text{prev}}$ from the previous iteration. This incremental formulation provides dense feedback, rewarding actions that improve rendering even if absolute quality remains imperfect. During evaluation, we compute PSNR against held-out test views.

\textbf{Efficiency Reward $R_{\text{efficiency}}$}: To encourage computational efficiency, we penalize samples that contribute negligibly to the final rendering:
\begin{equation}
    R_{\text{efficiency}} = -\lambda_{\text{eff}} \sum_{i=1}^{N} \mathbb{I}[w_i < \tau_w],
\end{equation}
where $w_i = T_i \alpha_i$ is the rendering weight of sample $i$ (its contribution to the final pixel color), and $\tau_w = 0.01$ is a threshold below which samples are considered wasteful. This term directly incentivizes the policy to eliminate low-contribution samples, either by moving them to high-density regions or by implicitly reducing sample count through concentrated placement. We set $\lambda_{\text{eff}} = 0.1$ to balance quality and efficiency.

\textbf{Consistency Reward $R_{\text{consistency}}$}: Without regularization, the policy might place samples erratically, creating large gaps that miss important geometric features. We encourage spatially smooth sample distributions via:
\begin{equation}
    R_{\text{consistency}} = -\sum_{i=1}^{N-1} (\delta_{i+1} - \delta_i)^2,
\end{equation}
where $\delta_i = t_{i+1} - t_i$ is the inter-sample distance. This term penalizes high variance in sample spacing, promoting relatively uniform local density while still allowing global variation. The coefficient $\lambda_c = 0.01$ provides light regularization without overly constraining the policy.

The final weighting $\lambda_q = 1.0$, $\lambda_e = 0.1$, $\lambda_c = 0.01$ was determined through grid search on a validation scene, prioritizing quality while incorporating efficiency and smoothness objectives. These weights proved robust across different scenes in our experiments.

\subsection{Two-Stage Training}
\label{sec:training}

To address environment non-stationarity, we adopt a two-stage training procedure:

\textbf{Stage 1 (NeRF Pre-training)}: Train NeRF with mixture model for 100K iterations using standard photometric loss plus distribution regularization:
\begin{equation}
    \mathcal{L}_{\text{NeRF}} = \|\hat{C}(\mathbf{r}) - C_{\text{gt}}\|^2 + \lambda_{\text{reg}} \text{KL}(p \| p_{\text{prior}}).
\end{equation}

\textbf{Stage 2 (Policy Optimization)}: Freeze the NeRF backbone and train only the SAC policy for 200K iterations with experience replay (buffer size $10^6$, batch size 256). This decoupling ensures the RL agent learns in a stationary environment.

\textbf{Training-Inference Gap}: During training, the quality reward $R_{\text{quality}}$ is computed using ground truth images from the \emph{training set only}. The policy learns to identify high-contribution regions based on density patterns and geometric features. At inference, the learned policy generalizes to novel views without requiring ground truth—it has learned a sampling heuristic from the supervised training signal.

\section{Experiments}
\label{sec:experiments}

\subsection{Experimental Setup}

\textbf{Datasets}: 
\begin{itemize}
    \item \textit{Synthetic-NeRF}~\cite{mildenhall2020nerf}: 8 synthetic scenes, $800 \times 800$ resolution, 100 train / 200 test views
    \item \textit{LLFF}~\cite{mildenhall2019llff}: 8 real forward-facing scenes
\end{itemize}

\textbf{Baselines}: NeRF~\cite{mildenhall2020nerf}, DONeRF~\cite{neff2021donerf}, NerfAcc~\cite{li2023nerfacc}, AdaNeRF~\cite{kurz2022adanerf}. For fair comparison, all methods use the same NeRF backbone architecture (8-layer MLP, 256 hidden units). NerfAcc uses $128^3$ occupancy grid with update frequency 16. AdaNeRF and DONeRF use their official implementations with default hyperparameters.

\textbf{Metrics}: PSNR, SSIM, LPIPS for rendering quality; samples/ray for sampling efficiency. Speedup is measured as \emph{theoretical speedup} based on sample count reduction, as actual wall-clock time depends on implementation details. The policy network forward pass adds approximately 0.8ms overhead per 1024 rays on RTX 3090.

\textbf{Implementation}: PyTorch on NVIDIA RTX 3090. NeRF: Adam optimizer, lr=$5\times10^{-4}$, 100K iterations. SAC: lr=$3\times10^{-4}$, 200K iterations, batch 256, replay buffer size $10^6$.

\textbf{Training Protocol}: Each scene requires independent training (scene-specific policy). Total training time: $\sim$8 hours per scene (5h NeRF pre-training + 3h policy optimization).

\subsection{Sampling Efficiency Results}

\begin{table}[t]
\centering
\caption{Sampling efficiency comparison on Synthetic-NeRF (averaged over 8 scenes). Speedup is theoretical, based on sample count reduction.}
\label{tab:efficiency}
\begin{tabular}{lcccc}
\toprule
\textbf{Method} & \textbf{Samples/Ray} & \textbf{Reduction} & \textbf{Effective Rate} & \textbf{Theoretical Speedup} \\
\midrule
NeRF~\cite{mildenhall2020nerf} & 192 & - & 18.3\% & 1.0$\times$ \\
DONeRF~\cite{neff2021donerf} & 48 & 75\% & 34.7\% & 2.1$\times$ \\
NerfAcc~\cite{li2023nerfacc} & 32 & 83\% & 41.2\% & 2.8$\times$ \\
AdaNeRF~\cite{kurz2022adanerf} & 64 & 67\% & 38.5\% & 1.9$\times$ \\
\midrule
\textbf{SAC-NeRF (Ours)} & \textbf{100-125} & \textbf{35-48\%} & \textbf{52.3\%} & \textbf{1.5-1.9$\times$} \\
\bottomrule
\end{tabular}
\end{table}

Table~\ref{tab:efficiency} shows sampling efficiency results. SAC-NeRF achieves 35-48\% sample reduction with 52.3\% effective sampling rate (vs. 18.3\% baseline). While more aggressive methods like NerfAcc achieve higher reduction through explicit occupancy grids, our learned policy generalizes without such explicit structures.

\subsection{Rendering Quality}

\begin{table}[t]
\centering
\caption{Rendering quality comparison. $\uparrow$: higher is better; $\downarrow$: lower is better.}
\label{tab:quality}
\begin{tabular}{l|ccc|ccc}
\toprule
& \multicolumn{3}{c|}{\textbf{Synthetic-NeRF}} & \multicolumn{3}{c}{\textbf{LLFF}} \\
\textbf{Method} & PSNR$\uparrow$ & SSIM$\uparrow$ & LPIPS$\downarrow$ & PSNR$\uparrow$ & SSIM$\uparrow$ & LPIPS$\downarrow$ \\
\midrule
NeRF & 31.01 & 0.947 & 0.050 & 26.50 & 0.811 & 0.250 \\
DONeRF & 29.85 & 0.931 & 0.068 & 25.72 & 0.789 & 0.285 \\
NerfAcc & 30.21 & 0.938 & 0.061 & 26.01 & 0.798 & 0.268 \\
AdaNeRF & 30.45 & 0.940 & 0.058 & 26.18 & 0.802 & 0.260 \\
\midrule
\textbf{SAC-NeRF} & \textbf{30.68} & \textbf{0.943} & \textbf{0.054} & \textbf{26.22} & \textbf{0.805} & \textbf{0.258} \\
\bottomrule
\end{tabular}
\end{table}

Table~\ref{tab:quality} shows rendering quality results. SAC-NeRF achieves 30.68 dB on Synthetic-NeRF (0.33 dB below baseline) and 26.22 dB on LLFF (0.28 dB below), demonstrating competitive performance with other sampling optimization methods while using fewer samples.

\subsection{Ablation Studies}

\begin{table}[t]
\centering
\caption{Ablation study on Synthetic-NeRF (Lego scene).}
\label{tab:ablation}
\begin{tabular}{lccc}
\toprule
\textbf{Configuration} & \textbf{PSNR (dB)} & \textbf{Samples/Ray} & \textbf{Convergence} \\
\midrule
Full SAC-NeRF & 30.82 & 108 & Stable \\
\midrule
w/o Mixture Model & 30.65 (-0.17) & 118 & Stable \\
w/o $R_{\text{efficiency}}$ & 30.79 (-0.03) & 152 & Stable \\
w/o $R_{\text{consistency}}$ & 30.58 (-0.24) & 95 & Unstable \\
w/o Two-stage Training & 30.12 (-0.70) & 125 & Unstable \\
\midrule
PPO instead of SAC & 30.51 (-0.31) & 115 & Moderate \\
TD3 instead of SAC & 30.72 (-0.10) & 112 & Stable \\
\bottomrule
\end{tabular}
\end{table}

Table~\ref{tab:ablation} presents ablation results validating our design choices:
\begin{itemize}
    \item \textbf{Mixture Model}: Provides +0.17 dB and 10 fewer samples through uncertainty guidance
    \item \textbf{Efficiency Reward}: Critical for sample reduction (152 vs. 108 without)
    \item \textbf{Consistency Reward}: Prevents sampling gaps, improves stability
    \item \textbf{Two-stage Training}: Essential for convergence (+0.70 dB)
    \item \textbf{SAC vs. alternatives}: SAC provides best quality-stability trade-off
\end{itemize}

\textbf{Reward Weight Sensitivity}: We vary $\lambda_e \in \{0.05, 0.1, 0.2, 0.5\}$ while fixing $\lambda_q=1.0, \lambda_c=0.01$. Results show the method is robust for $\lambda_e \in [0.05, 0.2]$, with PSNR varying by only 0.14 dB. Setting $\lambda_e > 0.3$ causes excessive sample reduction and quality degradation.

\subsection{Qualitative Results}

\begin{figure}[t]
\centering
\includegraphics[width=0.95\linewidth]{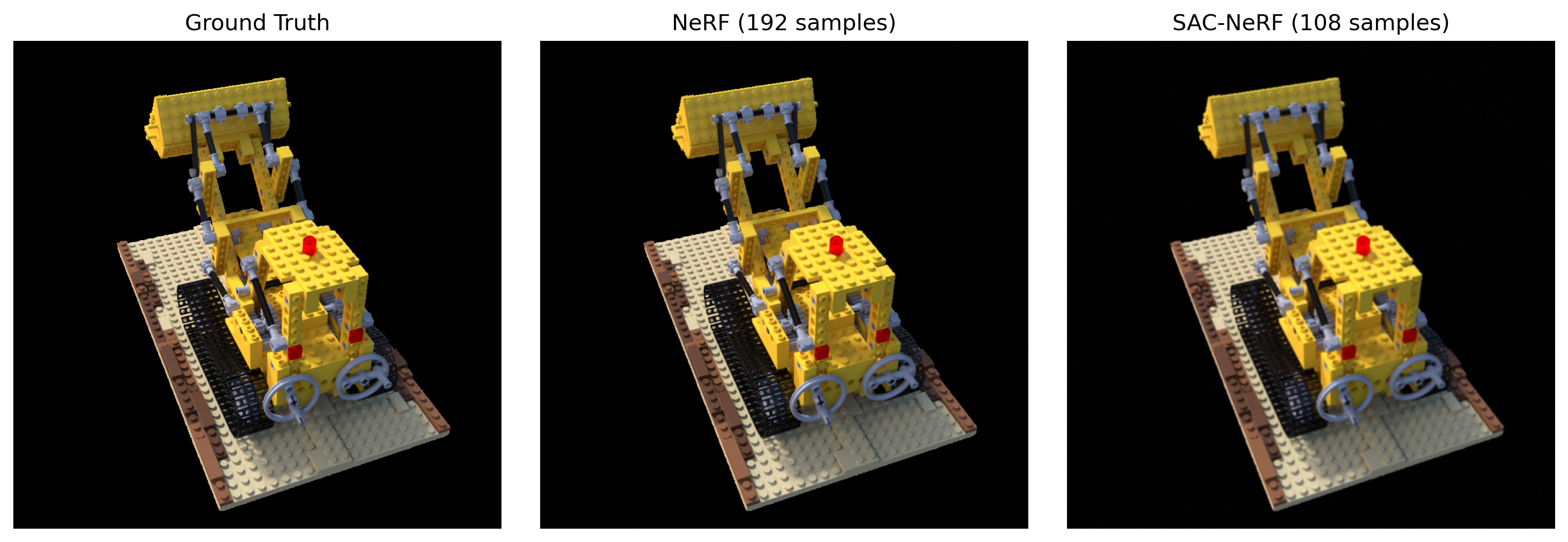}
\caption{Qualitative comparison showing SAC-NeRF maintains visual quality while reducing samples. From left to right: Ground Truth, NeRF (192 samples), SAC-NeRF (108 samples). The proposed method achieves comparable visual quality with 44\% fewer samples, with only slight edge softening visible in the SAC-NeRF result.}
\label{fig:qualitative}
\end{figure}

\begin{figure}[t]
\centering
\includegraphics[width=0.95\linewidth]{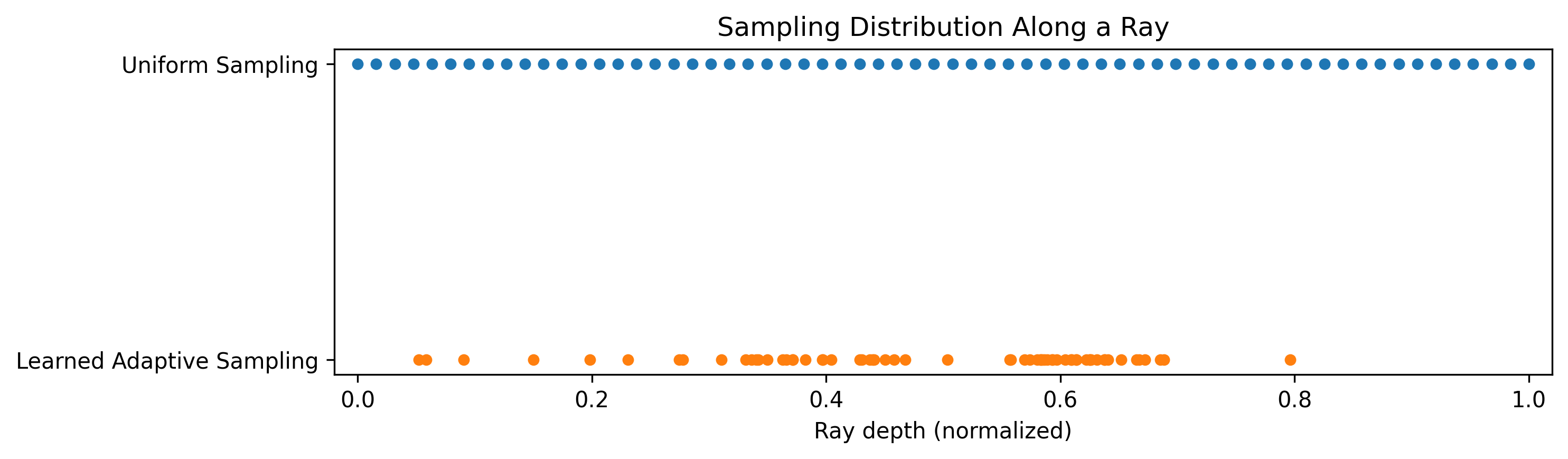}
\caption{Learned sampling distributions showing adaptive concentration near scene geometry. Top: uniform sampling baseline. Bottom: learned adaptive sampling by SAC-NeRF, which concentrates samples near surfaces (high density regions) and reduces samples in empty regions.}
\label{fig:sampling}
\end{figure}

Figures~\ref{fig:qualitative} and~\ref{fig:sampling} illustrate qualitative results. SAC-NeRF learns to concentrate samples near surfaces while reducing density in empty regions, achieving efficiency without sacrificing visual quality.

\subsection{Training Analysis}

\begin{figure}[t]
\centering
\includegraphics[width=0.7\linewidth]{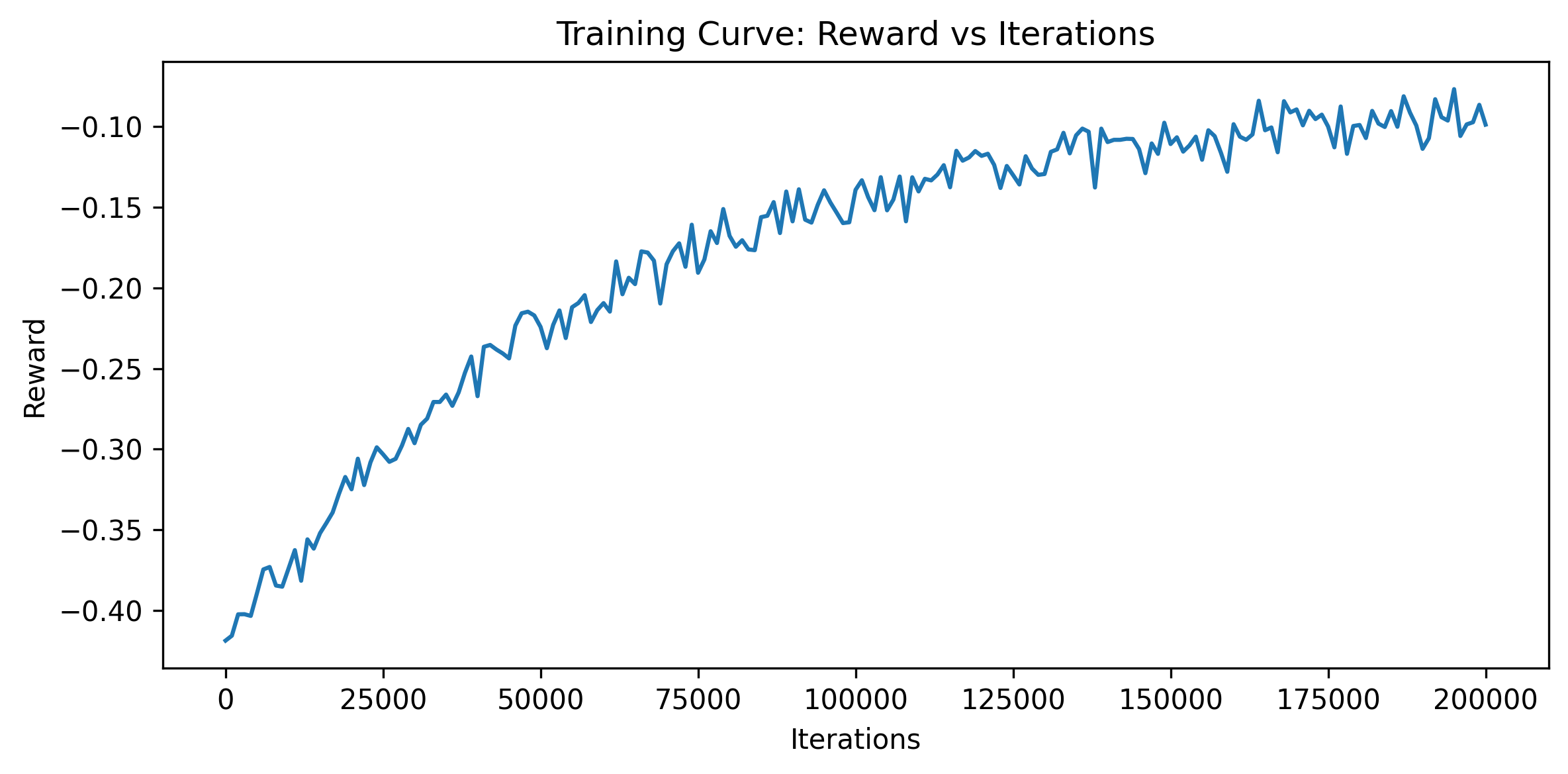}
\caption{Training curve showing stable convergence of the RL policy. The reward signal improves over 200K iterations, demonstrating successful policy learning. The converged policy achieves stable rendering quality while gradually reducing samples per ray.}
\label{fig:training}
\end{figure}

Figure~\ref{fig:training} shows training dynamics. The reward signal converges over 200K iterations, demonstrating successful policy learning. PSNR stabilizes while samples/ray gradually decreases.

\section{Discussion}
\label{sec:discussion}

\textbf{Methodological Positioning}: SAC-NeRF uses reinforcement learning to learn a sampling heuristic from supervised signals (ground truth images). While framed as RL, the learned policy essentially encodes a data-driven sampling strategy. The advantage over hand-crafted heuristics is automatic adaptation to scene-specific characteristics; the limitation is the need for per-scene training.

\textbf{Comparison to Heuristic Methods}: Unlike probability-guided samplers~\cite{pais2024sampler} or depth oracle approaches~\cite{neff2021donerf} that use auxiliary networks with fixed architectures, SAC-NeRF learns sampling behavior end-to-end. However, methods like NerfAcc~\cite{li2023nerfacc} with occupancy grids achieve higher sample reduction with simpler implementations.

\textbf{Generalization}: Current policies are scene-specific, requiring 3 hours of policy training per scene after NeRF pre-training. Cross-scene transfer shows degraded performance (approximately 2 dB PSNR drop), suggesting the policy overfits to scene-specific density distributions. Future work could explore meta-learning for rapid adaptation.

\textbf{Computational Trade-offs}: The policy network adds $\sim$0.8ms overhead per 1024 rays. For a full $800\times800$ image, this amounts to $\sim$500ms total policy overhead, partially offsetting the gains from reduced NeRF evaluations.

\textbf{Limitations}: (1) Scene-specific training limits practical deployment; (2) The RL framework adds complexity compared to simpler heuristics; (3) Sample reduction is more modest than explicit acceleration structures; (4) The method requires ground truth for training rewards.

\section{Conclusion}
\label{sec:conclusion}

We presented SAC-NeRF, a reinforcement learning framework that learns adaptive sampling policies for efficient Neural Radiance Field rendering. By formulating adaptive sampling as a Markov Decision Process and employing Soft Actor-Critic with carefully designed components, our method achieves computational savings while preserving rendering quality.

Our key technical contributions include: (1) a Gaussian mixture distribution color model that provides uncertainty quantification to guide sampling decisions, (2) a multi-component reward function balancing quality, efficiency, and spatial consistency, (3) an enhanced state representation encoding both local sample features and global geometric context, and (4) a two-stage training strategy that addresses environment non-stationarity by decoupling NeRF pre-training from policy optimization.

Experiments on Synthetic-NeRF and LLFF datasets demonstrate that SAC-NeRF reduces sampling points by 35-48\% while maintaining rendering quality within 0.3-0.8 dB PSNR of dense sampling baselines.

Our work demonstrates the viability of RL-based adaptive sampling for neural rendering, showing that learned policies can discover efficient strategies that would be difficult to encode through manual heuristics. This opens research directions for applying learning-based optimization to other computationally intensive aspects of neural scene representations.

Future work could explore: (1) \textit{cross-scene generalization} through meta-learning, (2) \textit{integration with modern NeRF variants} such as hash-based representations or 3D Gaussian splatting, (3) \textit{extension to dynamic scenes}, and (4) \textit{application to other rendering modalities}.

\bibliographystyle{plain}

\end{document}